\documentclass[final]{cvpr}

\usepackage{times}
\usepackage{epsfig}
\usepackage{graphicx}
\usepackage{amsmath}
\usepackage{amssymb}

\usepackage{tabularx}
\usepackage{lineno,hyperref}
\usepackage{times}
\usepackage{epsfig}
\usepackage{graphicx}
\usepackage{amsmath}
\usepackage{amssymb}
\usepackage{booktabs}
\usepackage{multirow}
\usepackage{subcaption}
\usepackage[export]{adjustbox}
\usepackage{float}
\usepackage{dashrule}
\usepackage{arydshln}
\usepackage[utf8x]{inputenc}
\usepackage[space]{grffile}
\usepackage{cleveref}

\usepackage{array, cellspace, makecell, caption}

\def\cvprPaperID{****} 
\def\confYear{CVPR 2021}

\begin{document}

\title{HPRNet: Hierarchical Point Regression 
for Whole-Body Human Pose Estimation
}

\author{Nermin Samet, Emre Akbas \\
Department of Computer Engineering \\
METU \\
{\tt\small \{nermin,emre\}@ceng.metu.edu.tr}
}

\maketitle

\begin{abstract}
In this paper, we present a new bottom-up one-stage method for whole-body  pose estimation, which we call ``hierarchical point regression,’’ or HPRNet for short. In standard body pose estimation, the locations of $\sim 17$ major joints on the human body are estimated. Differently, in whole-body pose estimation, the locations of fine-grained keypoints (68 on face, 21 on each hand and 3 on each foot) are estimated as well, which creates a scale variance problem that needs to be addressed.  To handle the scale variance among different body parts, we build a hierarchical point representation of body parts and jointly regress them. The relative locations of fine-grained keypoints in each part (e.g. face) are regressed in reference to the center of that part, whose location itself is estimated relative to the person center.  In addition, unlike the existing two-stage methods, our method  predicts whole-body pose in a constant time independent of the number of people in an image. On the COCO WholeBody dataset, HPRNet significantly outperforms all previous bottom-up methods on the keypoint detection of all whole-body parts (i.e. body, foot, face and hand); it also achieves state-of-the-art results on face (75.4 AP) and hand (50.4 AP) keypoint detection. Code and models are available at \url{https://github.com/nerminsamet/HPRNet}.
\end{abstract}


\section{Introduction}
 
As a challenging computer vision task, human pose estimation aims to localize human body keypoints in images and videos. Human pose estimation has an important role in several vision tasks and applications such as action recognition~\cite{du2015hierarchical, li2019actional, yan2019pa3d, huang2019part, luvizon20182d}, human mesh recovery~\cite{pose2mesh, kundu2020appearance, kama, kanazawa2018end}, augmented/virtual reality~\cite{cimen2018ar, elhayek2018fully, xu2019mo}, animation and gaming~\cite{azure1, azure2, huwai, animepose}. Unlike the standard human pose estimation task,   whole-body pose estimation aims to detect face, hand and foot keypoints in addition to the standard  human body keypoints. The challenge in this problem is the extreme  scale variance or imbalance among  different whole-body parts. For example, the relatively small scale of face and hand keypoints make accurate localization of face and hand keypoints more difficult compared to the standard body keypoints such as elbow, knee and hip. Direct application of existing human pose estimation methods do not yield satisfactory results due to this scale variance problem. 

Even though human pose estimation has been  well studied for the past few decades,  the whole-body pose estimation task has not been sufficiently explored, mainly due to the lack of large-scale fully annotated whole-body keypoint datasets. The previous few methods~\cite{openpose, sn}, trained several deep networks separately on different face, hand and body datasets, and ensembled them during inference. These methods suffer from issues arising from datasets’ biases, variations of illumination, pose and scales, and complex training and inference pipelines.

Recently, in order to address the missing benchmark issue, Jin \etal~\cite{wholebody}  introduced a  novel dataset for whole-body pose estimation, called COCO WholeBody. COCO WholeBody extends COCO keypoints dataset~\cite{coco} by further annotating face, hand and foot keypoints.  In addition to the standard, 17 human body keypoints from the COCO keypoints dataset; 68 facial landmarks, 42 hand keypoints and 6 foot keypoints are annotated (Figure~\ref{fig:wholebody_annots}). Along with these 133 whole-body keypoint annotations, the dataset also has face and hand bounding box annotations that were automatically computed from the extreme keypoints of the corresponding part. They   also proposed a strong baseline, called ZoomNet, which has set the state of the art. ZoomNet is a top-down, two-stage method based on the human pose estimation model HRNet~\cite{hrnetv2}. Given an image, ZoomNet first detects person instances using the FasterRCNN~\cite{faster} person detector, then it predicts 17 body and 6 foot keypoints using a CNN model. Later, to overcome the scale variance between whole-body parts, ZoomNet crops the hand and face areas that it detected and transforms them to higher resolutions using seperate CNNs to further perform face and hand keypoint estimation.

There are two main approaches for human pose and whole-body pose estimation; \textit{bottom-up}~\cite{paf, ning2017knowledge, ae, newell2016stacked, multiposenet, papandreou2018personlab, bulat2016human, pishchulin2016deepcut, sn, insafutdinov2017arttrack, insafutdinov2016deepercut, iqbal2017posetrack, jin2019multi, jin2017towards} and \textit{top-down}~\cite{chen2018cascaded, papandreou2017towards, mask, fang2017rmpe, hrnetv2, simple}.  Bottom-up methods directly detect human body keypoints and later group them to obtain final poses for each person in a given image. On the other hand, top-down methods (e.g. ZoomNet) first detect and extract person instances, then apply pose estimation on each instance separately. The grouping stage of bottom-up methods is more efficient than repeating pose estimation for each person instance. As a result, top-down methods slow down with the increasing number of people (Figure~\ref{fig:run_time}). However, compared to bottom-up methods, better accuracies are obtained by top-down approaches. 

In this paper, we propose a new bottom-up method, HPRNet, that explicitly handles the hierarchical nature of whole-body pose estimation by regressing keypoints hierarchically. To this end, in addition to the estimation of standard body keypoints, we define the bounding box centers of relatively small body parts such as face and hands with offsets to the person instance center (Figure~\ref{fig:kps}). Concurrently, we build another level of regression where we define each hand and face keypoints with an offset to their corresponding hand and face bounding box centers. We jointly train each level of regression hierarchy and regress all whole-body keypoints with respect to their defined center points. This hierarchical bottom-up approach brings two benefits. First, the scale variance among different body parts are handled naturally as the relative distances within each part are in a similar range and each part-type is processed by a separate sub-network. Second, being a bottom-up method, HPRNet’s inference speed is minimally affected by the number of persons in the input image. This is in contrast to the top-down methods such as ZoomNet, which significantly slows down with more person instances (65.7 ms for an image containing 1 person versus 668.2 ms for an image with 10 persons). Our method is based on the center-point based bottom-up object detection methods~\cite{centernet, houghnet, centernet2, cornernet}. These methods can easily be extended to the keypoint estimation task~\cite{centernet, houghnetjournal}. 

We validated the effectiveness of our method through ablation experiments and comparisons with the state of the art (SOTA) on the COCO WholeBody dataset. Our method significantly outperforms all bottom-up methods. It also outperforms the SOTA top-down method ZoomNet in the detection of face and hand keypoints, while being significantly faster than ZoomNet.

Our major contribution in this paper is the proposal of a one-stage,  bottom-up method to close the performance gap between the bottom-up and top-down methods. In contrast to top-down methods, our method runs almost in constant time, independent from the number of persons  in the input image.

\section{Related Work}
 
\subsection{Human Body Pose Estimation}

We can categorize the current approaches for multi-person pose estimation into two: bottom-up and top-down. In the bottom-up methods~\cite{paf, ning2017knowledge, ae, newell2016stacked, multiposenet, papandreou2018personlab, bulat2016human, pishchulin2016deepcut, sn, insafutdinov2017arttrack, insafutdinov2016deepercut, iqbal2017posetrack, jin2019multi, jin2017towards}, given an image, body keypoints detected first, without knowing the number or location of person instances or to which person instances these keypoints belong. Later, detected keypoints are grouped and assigned to person instances.
Recently, center-based object detection methods \cite{centernet} have been extended to perform human pose estimation~\cite{centernet, houghnetjournal}. These methods represent keypoints with an offset value to the center of the person box and directly regresses them during training. In order to improve localization of keypoints, they also estimate the heatmap of each keypoint as in other bottom-up methods~\cite{openpose,ae,papandreou2018personlab, multiposenet}. At inference, using center offsets, they group and assign keypoints to person instances. Since bottom-up methods detect all people keypoints at once, they are fast. 

Top-down methods~\cite{chen2018cascaded, papandreou2017towards, mask, fang2017rmpe, hrnetv2, simple} first detect person instances  in the input  image. Commonly, they use an off-the-shelf object detector (e.g. FasterRCNN~\cite{faster}) to obtain person boxes. Next, top-down methods estimate a single person pose for each cropped person box. By cropping and resizing each person box, top-down methods have the advantage to zoom into the details of each person. Therefore, top-down approaches are more capable of handling scale variance issues. As a result, state-of-the-art results are obtained by top-down methods and there is an accuracy gap between top-down and bottom-up approaches.  However, since a pose estimation model is run for each person instance, top-down methods tend to be slow on average, that is, they get significantly slower with increasing number of persons in an image (Figure \ref{fig:run_time}).

One may think that using human body pose estimation methods on a  whole-body pose estimation dataset (i.e. COCO WholeBody) could  be a solution for whole-body pose estimation. However, as it is stated in the COCO WholeBody dataset paper~\cite{wholebody},  due to the large scale variance between whole body parts, applying these methods directly results in suboptimal accuracies.

\begin{figure*}
\centering
\includegraphics[width=0.5\linewidth]{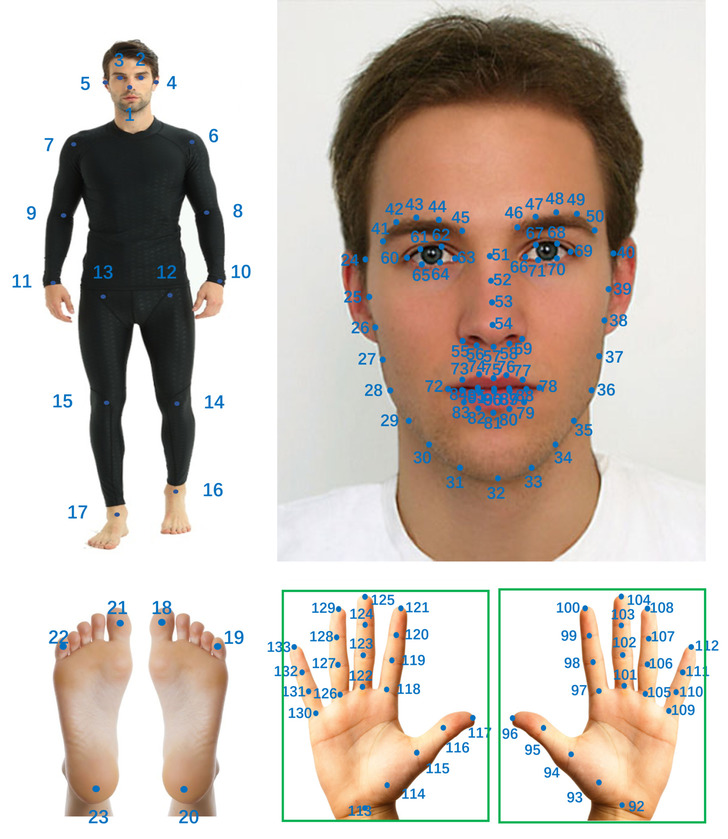}
\caption{Whole-body keypoints as defined in the COCO WholeBody dataset. There is a total of 133 keypoints. In addition to standard 17 human body keypoints (top-left) from the COCO keypoints dataset, there are 68 face (top-right), 42 hand (21 keypoints for each) (bottom-right) and 6 foot (3 for each) (bottom-left) keypoints are annotated. Image source: https://github.com/jin-s13/COCO-WholeBody}
\label{fig:wholebody_annots}
\end{figure*}

\subsection{Whole-body Pose Estimation}

Whole-body pose estimation requires accurate localization of keypoints on body, face, hand and feet. Detection of keypoints is well studied for each of these body parts independently, under face alignment~\cite{cao2014face, tzimiropoulos2015project, trigeorgis2016mnemonic, zhang2015learning}, facial landmark detection~\cite{retinaface, mtcnn},  hand pose estimation~\cite{oberweger2015hands, deepp}, hand tracking~\cite{sharp2015accurate, sridhar2015fast} and feet keypoint detection~\cite{openpose} topics.  However, there are not many works on the whole-body pose estimation mostly due to lack of a large-scale annotated dataset. Prior to the release of the COCO WholeBody dataset~\cite{wholebody}, OpenPose~\cite{openpose} attempted to detect the whole-body keypoints. For this purpose, OpenPose ensembles 5 separately trained models namely human body pose estimation, hand detection, face detection, hand pose estimation and face pose estimation. Due to these multiple models, training and inference of OpenPose are complex and costly. Our end-to-end trainable single network eliminates these drawbacks.

Hidalgo \etal~presented a bottom-up method called SN~\cite{sn}. Their model extends PAF~\cite{paf} for whole-body pose estimation. Similar to PAF~\cite{paf}, they predict heatmaps for each keypoint and use part affinity maps for grouping. SN model is trained on a dataset that is sampled from different datasets. Both SN and our proposed model HPRNet are bottom-up methods. However, SN falls short of handling scale variations between whole-body parts whereas hierarchical point representation of HPRNet overcomes this issue. 

\begin{figure*}
\centering
\includegraphics[width=1.0\textwidth]{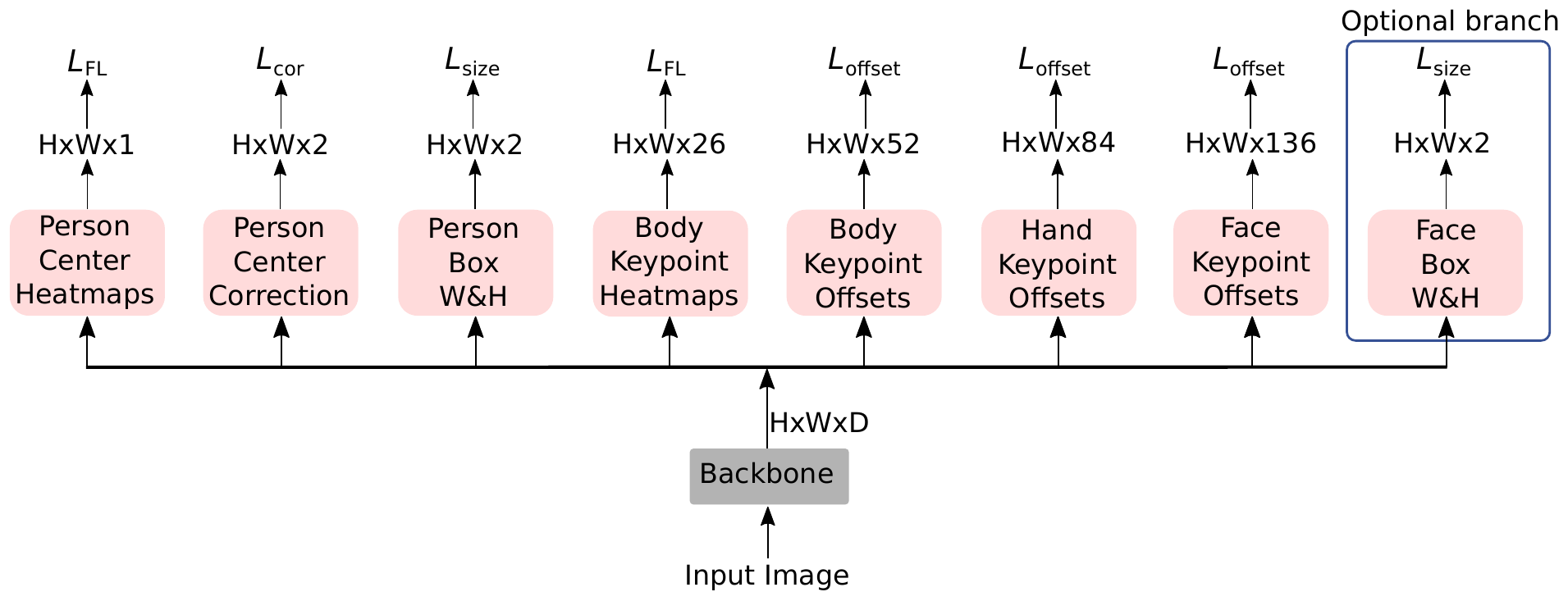}
\caption{Network architecture of the proposed HPRNet for whole-body keypoint detection.}
\label{fig:model}
\end{figure*}

The first step towards having a whole-body pose estimation benchmark is the release of the COCO WholeBody dataset~\cite{wholebody}. Jin \etal~extended the existing COCO keypoints~\cite{coco} dataset by further annotating face, hands and feet keypoints (Figure~\ref{fig:wholebody_annots}). They also proposed a strong, two-stage, top-down model to perform whole-body pose estimation on the COCO WholeBody dataset. Similar to top-down human pose estimation methods, Jin \etal~\cite{wholebody} first obtain candidate person boxes in an image using FasterRCNN~\cite{faster}.  Next, using a single network called ZoomNet, detection of whole-body keypoints is performed on the person boxes. ZoomNet is composed of 4 sub CNN networks. First, FeatureNet processes input person boxes and  extracts shared features at two scales. Next, using features from FeatureNet, BodyNet detects body and foot keypoints.  BodyNet is also responsible for the prediction of the face and hand bounding box corner points to roughly extract face and hand areas. Later, cropped face and hand bounding boxes are fed to the FaceHead and HandHead networks to detect the keypoints on face and hands. They use HRNet-W32~\cite{hrnetv2} network for the BodyNet and HRNetV2p-W18~\cite{hrnetw18} network for the FaceHead and HandHead networks.

Even though bottom-up approaches are fast, they are not robust enough to handle the scale variance across the whole-body parts. However, we hypothesize that representing each keypoint with an offset value to a carefully selected location can handle the scale variance. Based on this, we extend the center-based human pose estimation method~\cite{centernet} to perform whole-body pose estimation by introducing hierarchical regression of keypoints. We also show that hierarchical regression of keypoints for small scale whole-body parts (i.e. face and hand) is more effective than  cropping and zooming into them.

\section{Model}

HPRNet is a one-stage end-to-end trainable network that learns regressing the whole-body keypoints. In HPRNet, the input image first passes through a backbone network and output of the backbone is fed to 8 separate branches, namely;  \textit{Person Center Heatmap}, \textit{Person Center Correction}, \textit{Person W \& H},  \textit{Body Keypoint Offsets}, \textit{Body Keypoint Heatmaps}, \textit{Hand Keypoint Offsets}, \textit{Face Keypoint Offsets} and \textit{Face Box  W \& H}. We show the network architecture of HPRNet in Figure~\ref{fig:model}.

\subsection{Hierarchical Regression of Whole-Body Keypoints}

In HPRNet, we build a hierarchical regression mechanism, where we define each of the whole-body keypoints with a relative location (i.e. offset)  to a specific point on the person box. 

We represent each of the (standard) 17 keypoints on the body with an offset to the center of the person bounding box. Unlike the body; face, hand and foot are small parts. Based on this, we define each of this parts with a relative location to their part center as follows; (i) each of 68 face keypoints is defined with an offset to the center of face bounding box, (ii) each of 21 left hand keypoints is defined with an offset to left hand bounding box center, (iii) each of 21 right hand keypoints is defined with an offset to right hand bounding box center, (iv) each of 3 left foot keypoints is defined with an offset to left foot bounding box center, (v) each of 3 right foot keypoints is defined with an offset to right foot bounding box center. Face, hand and foot bounding boxes are automatically extracted from the groundtruth keypoint annotations.

\begin{figure}
\centering
\includegraphics[width=0.5\columnwidth]{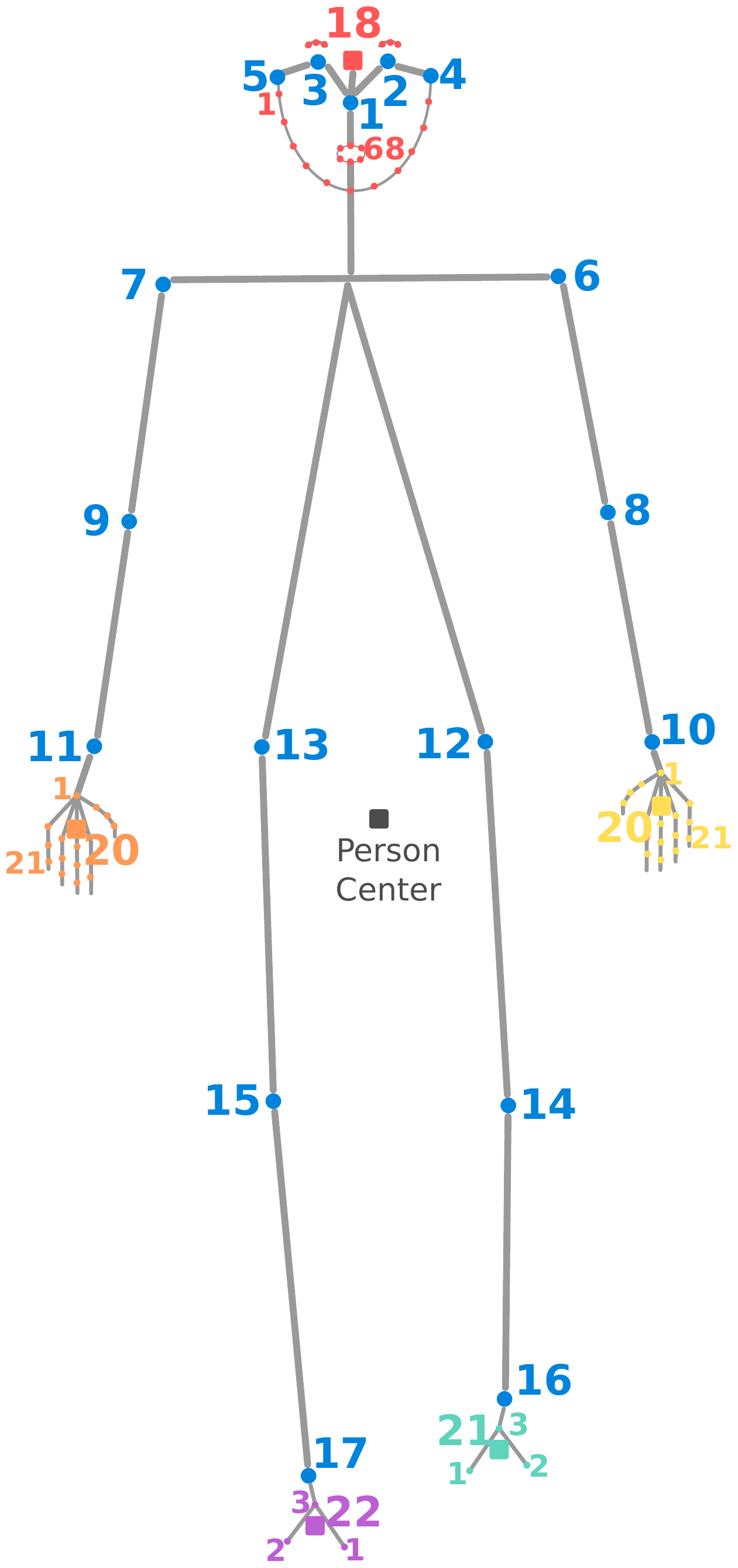}
\caption{All regressed keypoints in HPRNet. Blue keypoints are body keypoints as defined in COCO keypoints and COCO WholeBody datasets. Colored square points correspond to the face (18), left hand (19), right hand (20), left foot (21) and right foot (22) box centers. Blue keypoints (1-17) and colored square points are  defined with an offset to the center of the person instance. For simplicity, face and hand keypoints are sparsely illustrated. }
\label{fig:kps}
\end{figure}

We treat the bounding box center of the face, left hand, right hand, left foot and right foot as a body part keypoint and define each of them with an offset value to the person box center (Figure~\ref{fig:kps}). We illustrate the hierarchical regression of whole-body keypoints in~\Cref{fig:hier_v1}.

At inference, after detecting all the keypoints in the input image, we group and assign them  to person instances. To achieve this, we get predicted person centers from the output of \textit{Person Center Heatmap} branch as in CenterNet~\cite{centernet}. Next, we obtain the offset values on the predicted person center locations of \textit{Body Keypoint Offsets} branch output. After that, we add these offsets to person centers to obtain the \textit{regressed} body keypoint locations. At the same time, we extract the \textit{detected} body keypoints from the outputted heatmap of the \textit{Body Keypoint Heatmaps} branch. At the last step, we match the \textit{detected} and \textit {regressed} keypoints based on L2 distance and only take the keypoints inside the predicted person bounding box.

Next, we group face and hand keypoints (and foot keypoints as well, if we are using the Hierarchical Model-I (\Cref{fig:hier_v1}). We obtain  predicted part centers from the output of \textit{Body Keypoint Heatmaps} branch. Then, we collect  the offset values on the corresponding predicted part center locations of \textit{Hand Keypoint Offsets} and \textit{Face Keypoint Offsets} branch output. Finally, we add these offsets to the part centers to obtain the face and hand keypoints.

\subsection{Regression of Foot Keypoints}

\begin{figure*}
\centering
\begin{subfigure}{0.9\linewidth}
  \centering
  \includegraphics[width = \linewidth]{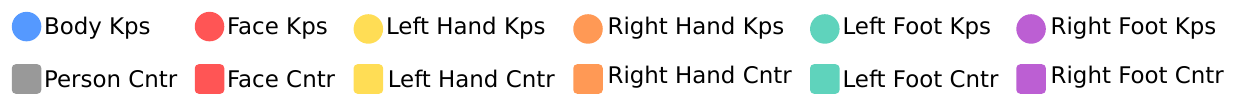}
\label{fig:hier_colors}
\end{subfigure}%

\begin{subfigure}[t]{.26\linewidth}
  \centering
  \includegraphics[width = \linewidth]{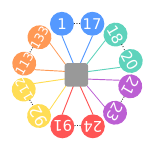} 
 \caption{Baseline Model}
\label{fig:hier_baseline}
\end{subfigure}%
\hspace{1em}
\begin{subfigure}[t]{.36\linewidth}
  \centering
  \includegraphics[width = \linewidth]{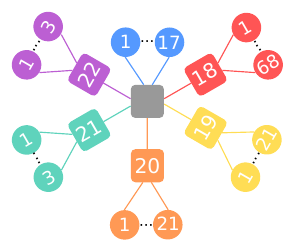}
\caption{Hierarchical Model-I}
\label{fig:hier_v1}
\end{subfigure}%
\hspace{1em}
\begin{subfigure}[t]{.29\linewidth}
  \centering
  \includegraphics[width = \linewidth]{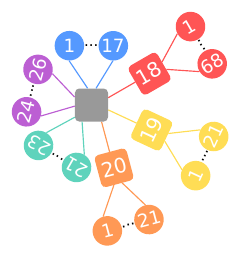}
\caption{Hierarchical Model-II}
\label{fig:hier_v2}
\end{subfigure}%

\caption{Hierarchical representations of whole-body keypoints. (a) Each of 133 whole-body keypoints is defined with an offset to the person box center. (b) Body keypoints and other part centers (i.e. foot, face and hand) are defined with offsets to the person box center. Foot, face and hand keypoints are defined with offsets
according to their corresponding part centers.  (c) Considering the sparsity of foot keypoint annotations, we define them with their offset values to the person box center. In both Hierarchical Model-I and Hierarchical Model-II, body keypoints are defined with offset values
 to the person box center. Each face and hand keypoint is defined with an offset to face and hand bounding box centers, respectively.}
\label{fig:hier_models}
\end{figure*}

Ideally, each labeled foot part in the COCO WholeBody dataset should have 3 keypoint annotations. However, more than 20\% of annotated feet  have missing annotations (i.e. they have one or two keypoints annotations, instead of three). These missing annotations present a challenge to HPRNet, since we automatically extract foot centers from the annotated extreme points. In the case of the missing foot keypoints, the obtained foot center point is not reliable. To deal with this issue, we treat the foot keypoints as body keypoints as shown in~\Cref{fig:hier_v2}, and represent them by their offsets to the center of the person bounding box.

\subsection{Network Architecture}
 
Given an input image of size $4H \times 4W \times 3$,  the backbone network outputs a feature map of size $H \times W \times D$. The backbone’s output is fed to the following subsequent branches. Each branch has one convolutional layer with $3 \times 3$ filters followed by a ReLU layer and another convolutional layer with $1 \times 1$ filters.

\begin{itemize}
\setlength{\itemsep}{0pt plus 0pt}

\item \textit{Person Center Heatmap} branch outputs $H \times W$ sized tensor for person center point predictions.
\item \textit{Person Center Correction} branch predicts $H \times W \times 2$  sized  tensor for the local offsets of center locations across the spatial axes. These offsets help to recover the lost precision of the center points due to down-sampling operations through the network.
\item \textit{Person Box W \& H} branch outputs  $H \times W \times 2$ sized tensor of widths and heights for each person instance center.
\item \textit{Body Keypoint Offsets} branch predicts offset values of 26 keypoints (17 body keypoints + 6 foot keypoints + Center of Face Box + Center of Left Hand Box + Center of Right Hand Box) to the person box center across the $x$ and $y$ axes.
\item \textit{Body Keypoint Heatmaps} branch outputs $H \times W \times 26$  sized heatmap tensor for the 26 keypoints.
\item \textit{Hand Keypoint Offsets} branch outputs  $H \times W \times 84$ sized tensor of offset values between 21 left hand keypoints and left hand box center; and the offset values between the 21 right hand keypoints and right hand box center across the spatial axes.
\item \textit{Face Keypoint Offsets} branch outputs  $H \times W \times 136$ sized tensor of  offset values between 68 face keypoints and face box center across the spatial axes.
\item \textit{Face Box  W \& H} branch outputs $H \times W \times 2$ sized tensor of widths and heights for each face. It is an optional branch. 
\end{itemize}

\begin{table*}
\caption{Comparison of Hierarchical Model-I (HM-I) and Hierarchical Model-II (HM-II) as in  Figure~\ref{fig:hier_models}. Training foot keypoints with offset values to the person box center outperforms the model when trained with offset values to the foot part centers. Both models are trained with DLA backbone. }
\centering
\resizebox{0.9\textwidth}{!}{
\begin{tabular}{lcccccccccccc}
\toprule
\multirow{2}{*}{Method} & \multicolumn{2}{c}{body}  & \multicolumn{2}{c}{foot} & \multicolumn{2}{c}{face} & \multicolumn{2}{c}{hand} & \multicolumn{2}{c}{whole-body} & \multicolumn{2}{c}{all-mean} \\ 
\cmidrule(lr){2-3}  \cmidrule(lr){4-5} \cmidrule(lr){6-7} \cmidrule(lr){8-9} \cmidrule(lr){10-11} \cmidrule(lr){12-13}
               &       \textit{AP$^{kp}$} & \textit{AR$^{kp}$} &  \textit{AP$^{kp}$} & \textit{AR$^{kp}$}  & \textit{AP$^{kp}$} & \textit{AR$^{kp}$} & \textit{AP$^{kp}$} & \textit{AR$^{kp}$}  & \textit{AP$^{kp}$} & \textit{AR$^{kp}$}   & \textit{AP$^{kp}$} & \textit{AR$^{kp}$}\\ \hline
HM-I  & \textbf{55.5}  &  \textbf{63.4}  & 33.5 & 55.3 & \textbf{74.6} & 83.5 & 44.1 & 57.8 & 28.0 & 40.5 & 47.1 & 60.1  \\ 
HM-II &  55.2  & 63.1 & \textbf{49.1} &  \textbf{60.9} & \textbf{74.6} &  \textbf{83.7} & \textbf{47.0}  & \textbf{60.8} & \textbf{31.5} & \textbf{44.6} & \textbf{51.5}& \textbf{62.6}  \\ 
\bottomrule
\end{tabular}}%
\label{table:kp_foot}
\end{table*}

\begin{table*}
\caption{Comparing HPRNet with the baseline model. To obtain the baseline results, we regress all the 133 keypoints to the person instance box center during training as in ~\Cref{fig:hier_v1}. Both models are trained with DLA backbone.}
\centering
\resizebox{0.9\textwidth}{!}{
\begin{tabular}{lcccccccccccc}
\toprule
\multirow{2}{*}{Method} & \multicolumn{2}{c}{body}  & \multicolumn{2}{c}{foot} & \multicolumn{2}{c}{face} & \multicolumn{2}{c}{hand} & \multicolumn{2}{c}{whole-body} & \multicolumn{2}{c}{all-mean} \\
\cmidrule(lr){2-3}  \cmidrule(lr){4-5} \cmidrule(lr){6-7} \cmidrule(lr){8-9} \cmidrule(lr){10-11} \cmidrule(lr){12-13}
                 &       \textit{AP$^{kp}$} & \textit{AR$^{kp}$} &  \textit{AP$^{kp}$} & \textit{AR$^{kp}$}  & \textit{AP$^{kp}$} & \textit{AR$^{kp}$} & \textit{AP$^{kp}$} & \textit{AR$^{kp}$}  & \textit{AP$^{kp}$} & \textit{AR$^{kp}$} & \textit{AP$^{kp}$} & \textit{AR$^{kp}$}  \\ \hline
Baseline  &  46.7  &  55.5  & 33.6 & 48.9 & 52.0 & 60.2 & 26.4 & 39.1 & \textbf{33.3} & 43.4  & 38.4 &  49.4 \\ 
HPRNet &  \textbf{55.2}  & \textbf{63.1} & \textbf{49.1} &  \textbf{60.9} & \textbf{74.6} &  \textbf{83.7} & \textbf{47.0}  & \textbf{60.8} & 31.5 & \textbf{44.6} & \textbf{51.5}& \textbf{62.6}  \\ 
\bottomrule
\end{tabular}}%
\label{table:kp_baseline}
\end{table*}

\begin{table*}
\caption{Comparison with the state-of-the-art on COCO WholeBody validation set. The methods are divided into two groups: top-down and bottom-up. The best results and run times are boldfaced separately for each group. HPRNet performs best among the bottom-up methods.  HPRNet also obtains state-of-the-art results on face and hand keypoint detection outperforming ZoomNet. Among all methods, HPRNet with DLA backbone is the fastest one. $^*$ indicates that run time linearly increases as the number of people in an image increases. HG is Hourglass-104. R. time is Running time.}
\centering
\resizebox{1.0\textwidth}{!}{
\setlength{\tabcolsep}{8pt}
\begin{tabular}{lrlrlrlrlrlrlc}

\toprule
\multirow{2}{*}{Method} & \multicolumn{2}{c}{body}  & \multicolumn{2}{c}{foot} & \multicolumn{2}{c}{face} & \multicolumn{2}{c}{hand} & \multicolumn{2}{c}{whole-body} & \multicolumn{2}{c}{all-mean} & \multirow{2}{*}{\shortstack[c]{R. time \\ (ms)}} \\
\cmidrule(lr){2-3}  \cmidrule(lr){4-5} \cmidrule(lr){6-7} \cmidrule(lr){8-9} \cmidrule(lr){10-11} \cmidrule(lr){12-13}
&       \textit{AP$^{kp}$} & \textit{AR$^{kp}$} &  \textit{AP$^{kp}$} & \textit{AR$^{kp}$}  & \textit{AP$^{kp}$} & \textit{AR$^{kp}$} & \textit{AP$^{kp}$} & \textit{AR$^{kp}$}  & \textit{AP$^{kp}$} & \textit{AR$^{kp}$}  & \textit{AP$^{kp}$} & \textit{AR$^{kp}$}  &   \\ \hline
\midrule 

\textit{Top-down methods:} & & & & & & & & & & &\\ 
OpenPose~\cite{openpose} &  56.3  & 61.2  &  53.2  &  64.5  & 48.2 &  62.6 & 19.8  & 34.2 & 33.8 &  44.9  & 42.3 & 53.5 &  \textbf{45} \\ 
HRNet$^*$~\cite{hrnetv2} & 65.9 & 70.9 & 31.4 & 42.4 & 52.3 & 58.2 & 30.0 & 36.3 & 43.2 & 52.0 & 44.6 & 52.0 & - \\ 
ZoomNet$^*$~\cite{wholebody} & \textbf{74.3} & \textbf{80.2} & \textbf{79.8} & \textbf{86.9} & \textbf{62.3} & \textbf{70.1} & \textbf{40.1} & \textbf{49.8} & \textbf{54.1} &  \textbf{65.8} & \textbf{62.1} &  \textbf{70.6} & 175 \\ 
\midrule 
\textit{Bottom-up methods:}  \\ 
PAF~\cite{paf} & 26.6 & 32.8 & 10.0 & 25.7 & 30.9 & 36.2 & 13.3 & 32.1 & 14.1 & 18.5 & 19.0 &  29.1 & 100 \\ 
SN~\cite{sn}  & 28.0 & 33.6 & 12.1 & 27.7 & 38.2 & 44.0 & 13.8 & 33.6 & 16.1 & 20.9 & 21.6 & 32.0 & 216 \\ 
AE~\cite{ae} & 40.5 & 46.4 & 7.7 & 16.0 & 47.7 & 58.0 & 34.1 & 43.5 & 27.4 & 35.0 & 31.5 &  39.8 & -\\ 
Ours (HPRNet-DLA) &  55.2  & 63.1  &  49.1  &  60.9  & 74.6  & 83.7 & 47.0   & 60.8  & 31.5 &  44.6  & 51.5 & 62.6 & \textbf{37} \\ 
Ours (HPRNet-HG) &  \textbf{59.4}  & \textbf{68.3}  &  \textbf{53.0}  &  \textbf{65.4}  & \textbf{75.4}   & \textbf{86.8} &  \textbf{50.4}  &  \textbf{64.2} & \textbf{34.8} &  \textbf{49.2}  & \textbf{54.6} & \textbf{66.8} & 101 \\ 
\bottomrule
\end{tabular}}%
\label{table:sota}
\end{table*}

\subsection{Objective Functions}

For the optimization of the \textit{Person Center Heatmap (PCH)} and \textit{Body Keypoint Heatmap (BKH)} branches, we use the modified focal loss~\cite{retinanet} as done in previous work~\cite{cornernet, extremenet, centernet, houghnet}. Modified focal loss (FL) is presented in Equation~\ref{eq:focalloss}. $I \in R^{4W \times 4H \times 3}$ is our input image. In HPRNet,due to downsampling operations, the spatial output size of each branch is 4 times smaller resulting in $W \times H$. Therefore, $Y \in[0,1]^{W \times H \times C}$ is the ground truth heatmap for person centers and keypoints. $C$ corresponds to class number and keypoint types. For instance, in the \textit{Person Center Heatmap} branch, we have only \textit{person} class, thus $C=1$.  $\hat{Y} \in[0,1]^{W \times H \times C}$ is the predicted heatmap output by the branches where $\hat{Y}_{x, y, c}=1$ indicates presence of a person center or keypoint at location $(x,y)$ for class $c$. In the following all equations, $N$ is the total number of ground truth person centers or keypoints in image $I$. $\alpha$ and $\beta$  are focal loss parameters and set as  $\alpha=2$ and $\beta=4$ as in CornerNet~\cite{cornernet}. 

\begin{equation}  \label{eq:focalloss}
L_\mathrm{FL} =\frac{-1}{N} \sum_{x y c}\left\{\begin{array}{cl}
\left(1-\hat{Y}_{x y c}\right)^{\alpha} \log \left(\hat{Y}_{x y c}\right) & \text { if } Y_{x y c}=1 \\
\left(1-Y_{x y c}\right)^{\beta}\left(\hat{Y}_{x y c}\right)^{\alpha} & \\
\log \left(1-\hat{Y}_{x y c}\right) & \text { otherwise }
\end{array}\right.
\end{equation}

To compensate for the discretization error of the person center points due to down-sampling operations through the network, we optimize the \textit{Person Center Correction} according to the following L1 loss similar to the bottom-up object detectors~\cite{cornernet, extremenet, centernet, houghnet}. 
$\hat{T} \in \mathcal{R} ^ {W \times H \times 2}$ is the predicted local offset by the network to recover the lost precision of person center points. $p \in \mathcal{R}^{2}$ is a ground truth keypoint and $\tilde{p}=\left\lfloor\frac{p}{4}\right\rfloor$ is the corresponding ground keypoint location at low-resolution.

\begin{equation}  \label{eq:correction}
L_\mathrm{cor}=\frac{1}{N} \sum_{p}\left|\hat{T}_{\tilde{p}}-\left(\frac{p}{4}-\tilde{p}\right)\right|
\end{equation}

We optimize the \textit{Body Keypoint Offset}, \textit{Hand Keypoint Offset} and \textit{Face Keypoint Offset} branches using the L1 loss. The generic formulation of keypoint regression is presented in Equation~\ref{eq:offsetreg}. In the equation, $\hat{O} \in \mathbb{R}^{H \times W \times k \times 2}$ is the regression output of keypoints $k$ for a specific whole-body part (i.e. body, face, hand), and $B_\mathrm{part}$ is the ground truth center of that part’s bounding box.

\begin{equation}  \label{eq:offsetreg}
L_\mathrm{offset} = \sum_{k}\left|\hat{O}_{[k]} - B_\mathrm{part}\right|
\end{equation}

Finally, for the \textit{Person Box H \& W} and \textit{Face Box H \& W} branches, we use L1 loss and scale it  by 0.1 as in CenterNet~\cite{centernet}. In the Equation~\ref{eq:size}, $s_{n}=\left(w, h\right)$ is the width and height values of the each object (or face) $n$ and $\hat{S} \in \mathcal{R}^{W \times H \times 2}$ is the predicted width and height values.

\begin{equation}  \label{eq:size}
L_\mathrm{size}=\frac{1}{N} \sum_{n=1}^{N}\left|\hat{S}_{p_{n}}-s_{n}\right|
\end{equation}

We obtain the overall loss by summing the losses from all branches as follows:


\begin{equation} 
\begin{split}
L_\mathrm{overall}  = &  L_\mathrm{FL}^{PCH} + L_\mathrm{FL}^{BKH} + L_\mathrm{cor} + L_\mathrm{offset}^{body} + L_\mathrm{offset}^{face} \\ & + L_\mathrm{offset}^{hand} + 0.1 L_\mathrm{size}^{person} +  0.1 L_\mathrm{size}^{face}
\end{split}
\end{equation}

\begin{figure*}
\captionsetup[subfigure]{labelformat=empty}
\centering
\begin{subfigure}[b]{0.278\textwidth}
 \includegraphics[width=1.0\textwidth]{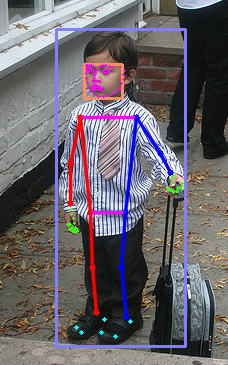}
 \includegraphics[width=1.0\textwidth]{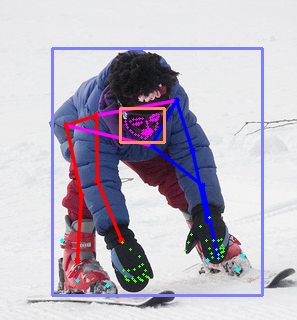}
\end{subfigure}
\begin{subfigure}[b]{0.3\textwidth}
 \includegraphics[width=1.0\textwidth]{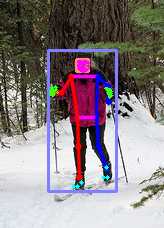}
 \includegraphics[width=1.0\textwidth]{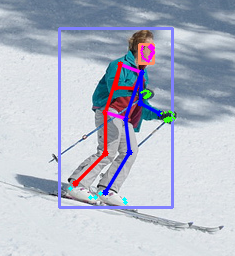}
\end{subfigure}
\begin{subfigure}[b]{0.3\textwidth}
 \includegraphics[width=1.0\textwidth]{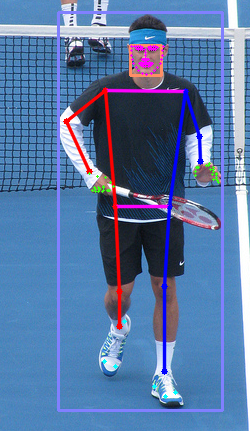}
 \includegraphics[width=1.0\textwidth]{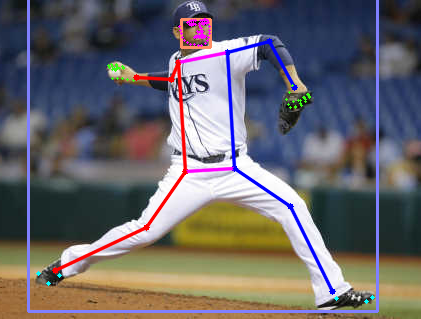}
\end{subfigure}
\caption{Sample whole-body keypoint detection results of HPRNet. We show correctly detected people, and their whole-body poses. Detection box is marked with a purple bounding box, and body pose estimation is shown with blue color for the left parts, red color for the right parts. For clarity, we mark the detected keypoints on face, hand and foot with magenta, green and cyan colors, respectively. Detected faces are marked with an orange bounding box.}
\label{fig:vis-results}
\end{figure*}

\section{Experiments}

This section describes the experiments we conducted to show the effectiveness of our proposed method. First, we present ablation experiments to compare hierarchical models I and II shown in Figure~\ref{fig:hier_models}.  Next, we compare our method with our baseline CenterNet~\cite{centernet} (\Cref{fig:hier_baseline}). Finally, we provide performance comparison with the state of the art and a run-time analysis.

\paragraph{\textbf{Implementation Details}}  We use Deep Layer Aggregation (DLA)~\cite{dla} backbone for ablation and baseline comparison experiments, and Hourglass-104~\cite{cornernet}  as our backbone network for  state of the art comparison. For all experiments, during training we resize the images to $512 \times 512$ pixels. At inference we use images with their original sizes without applying any scaling. We train all the models with a batch size of 32 for 140 epochs using the Adam optimizer~\cite{adam}. We set the initial learning rate to $1.25 \times 10^{-4}$ and divided it by 10 at epochs 90 and 120.  We trained all of the models on 4 Tesla V100 GPUs, and tested using a single GTX 1080 TI GPU. We used PyTorch~\cite{pytorch} to implement our models. All of our experiments are conducted on the COCO WholeBody Dataset~\cite{wholebody} and results are presented in keypoint AP (AP$^{kp}$) and keypoint recall AR (AR$^{kp}$) metrics without any test time augmentation. All results are obtained on the COCO WholeBody validation set.

\subsection{Hierarchical Model-I vs Hierarchical Model-II}

In Table~\ref{table:kp_foot}, we compare Hierarchical Model-I and Hierarchical Model-II (see Figure~\ref{fig:hier_models}). As it can be seen from the table, regressing foot keypoints as a part of the body keypoints, improves the foot AP$^{kp}$ significantly by 15.6 points (33.5 vs. 49.1). Moreover, hand and whole-body AP$^{kp}$s also improved about 3 points in this setup. Based on these results, for the rest of the experiments we use the Hierarchical Model-II.

\subsection{Comparison with Baseline}

To obtain the baseline results, we regress all the 133 keypoints to the person instance box center during training as in CenterNet~\cite{centernet} (see~\Cref{fig:hier_v1}). In Table~\ref{table:kp_baseline}, we compare HPRNet with the baseline model in terms of accuracy and recall. Our proposed HPRNet significantly outperforms the baseline results for all AP$^{kp}$ and AR$^{kp}$  metrics except the whole-body AP$^{kp}$.

\begin{table*}
\caption{Face detection results. The first group of results are obtained from extreme face keypoints for both ZoomNet and HPRNet. The HPRNet results in the second group are obtained with an extra face detection branch. HG is Hourglass-104. }
\centering
\resizebox{0.8\linewidth}{!}{
\begin{tabular}{lccccccc}
 \toprule 
 \textbf{Method} & \textit{AP}& \textit{AP$_{50}$} &  \textit{AP$_{75}$} & \textit{AP$_M$} & \textit{AP$_L$} \\
 \midrule 
\textit{When face boxes are extracted from extreme face keypoints:} & & & & & \\

  ZoomNet  & 37.7 & 64.5  & 41.1 & 25.8 & 44.9 	\\
  HPRNet (DLA) & 46.2 & 70.6  & 54.8 & 32.2 & 53.8 	\\
   HPRNet (HG)  	& 46.1 & 70.9  & 53.6 & 33.4 & 53.1 	\\
\midrule 
\textit{ Our model with an extra face detection branch} & & & & & \\

HPRNet (DLA) & 55.8 & 82.3  & 66.2 & 40.0 & \textbf{63.6} 	\\
HPRNet (HG) & \textbf{56.4} & \textbf{82.4} & \textbf{67.1} & \textbf{43.4} & 63.3 	\\
\bottomrule
\end{tabular} }
\label{table:face_detection_comp}
\end{table*}

\subsection{Comparison with the State-of-the-art}

Table~\ref{table:sota} presents the performance  of our models and  several established keypoint estimation models on the COCO WholeBody validation set. We also present average run times if available. \textbf{HPRNet performs best among the bottom-up methods.} Other bottom-up methods especially fail to accurately localize foot keypoints. The performance gap between the second best performing bottom-up method and our method is 40.9 AP$^{kp}$ points on the foot keypoint detection. Similarly, our method outperforms other bottom-up methods for the body, face, hand and whole-body keypoint detection by a large margin. Among the top-down methods, ZoomNet outperforms the well known OpenPose~\cite{openpose} and HRNet~\cite{hrnetv2}. Here, ZoomNet is a two-stage framework where at the first stage person candidates are extracted with FasterRCNN and at the second stage ZoomNet is run on these candidate boxes. HRNet can be seen as a one-stage counterpart of ZoomNet and finally OpenPose is a multi-model which requires separate training for each whole-body part.  HPRNet obtains state-of-the-art results on face and hand keypoint detection. Our model with Hourglass-104 backbone  outperforms ZoomNet on the detection of face keypoints by 13.1 AP$^{kp}$ points and hand keypoints by 10.3 AP$^{kp}$ points. These successful results on the face and hand keypoint detection, further shows the effectiveness of our proposed bottom-up hierarchical approach over the ZoomNet’s zoom-in mechanism. However, for the detection of the body and whole-body keypoints ZoomNet performs best among all methods. \textbf{Among all methods, our HPRNet with the DLA backbone is the fastest one (37 ms) with constant run time}. In Figure~\ref{fig:vis-results}, we show sample qualitative results for our approach.

\begin{figure}
\centering
\includegraphics[width=1.0\columnwidth]{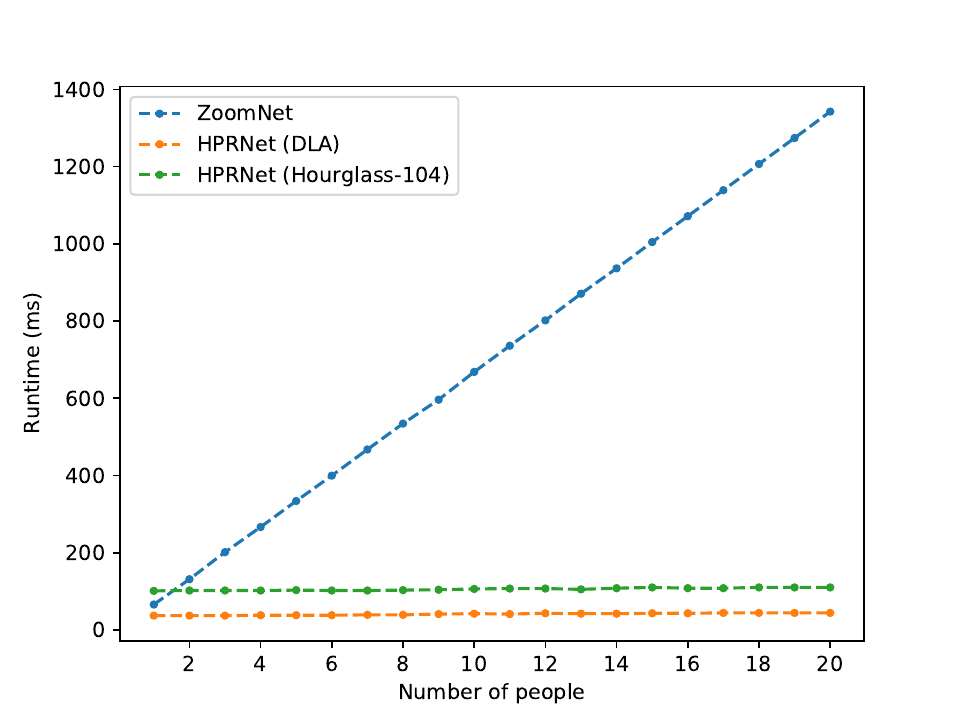}
\caption{Runtime analysis of ZoomNet and our models with respect to number of people in an image. As the number of people in an image increases, the runtime of ZoomNet linearly increases. Whereas, our models almost have constant run time.}
\label{fig:run_time}
\end{figure}

\paragraph{\textbf{Runtime Analysis}} Average run time of ZoomNet (including Faster RCNN for person detector) on a single image is 174.7 ms. Similarly, the average run time of HPRNet with DLA and Hourglass-104 backbones is 37 ms (26 ms for feedforward and 11 ms for keypoint grouping and assignment) and 101 ms (90 ms for feedforward and 11 ms for  keypoint grouping and assignment). HPRNet is significantly faster than ZoomNet. Moreover, as a top-down method, run time of ZoomNet increases as the number of people on an image increases. We compare the run time of our models and ZoomNet in Figure~\ref{fig:run_time}.

\subsection{Face Detection from Keypoints}

In this section, we studied the face detection task and compared HPRNet and ZoomNet.  We first extracted face boxes using extreme face keypoints and calculated AP scores as in object detection. Our model outperformed ZoomNet in face detection (46.2 AP vs 37.7 AP). Later, using an additional branch for face detection we train another model (see Figure~\ref{fig:model}). Our model with an extra face detection branch further improved the performance of HPRNet for face detection achieving 55.8 AP and 56.4 AP with DLA and Hourglass-104 backbones respectively. Results are presented in Table~\ref{table:face_detection_comp}.

 \section{Conclusion}

In this work, we introduced HPRNet as a bottom-up,  one-stage method for whole-body keypoint detection.  HPRNet handles scale variance among  whole-body parts by hierarchically regressing whole-body keypoints. We evaluated the effectiveness of our method through baseline comparison and ablation experiments on hierarchical structure of whole-body keypoints. Our method achieves state-of-the-art results in the detection of face and hand keypoints on the COCO WholeBody dataset; it also outperforms all other bottom-up methods in the detection of all whole-body parts. We conducted a run time analysis between HPRNet and ZoomNet and showed that in contrast to ZoomNet, HPRNet runs in constant time, independent of the number of persons in an image. 

\section{Acknowledgements}

The numerical calculations reported in this paper were fully performed at TUBITAK ULAKBIM, High Performance and Grid Computing Center (TRUBA resources).

{\small
\bibliographystyle{ieee_fullname}
\bibliography{refs}
}

\end{document}